\documentclass[letterpaper, 10 pt, conference]{ieeeconf}  

\IEEEoverridecommandlockouts

\usepackage{graphics} 
\usepackage{times} 
\usepackage{amsmath} 
\usepackage{amssymb}  
\usepackage{multirow}
\usepackage{booktabs}
\usepackage{relsize}
\usepackage{xspace}
\usepackage{bbold}
\usepackage{subfig}
\usepackage{listings}
\usepackage{bm}
\usepackage{bbm}
\usepackage{mathtools}
\usepackage{float}
\usepackage{textcomp}
\usepackage{gensymb}
\usepackage{tabularx}
\usepackage{makecell}
\usepackage{multirow}
\usepackage{hyperref}
\usepackage{booktabs}

\usepackage[color=todocolor, colorinlistoftodos]{todonotes}

\DeclareMathOperator{\se3}{\mathbf{SE(3)}}

\newcommand{\R}{\mathbb{R}}

\newcommand*{\C}[1]{\mathcal{#1}}
\newcommand*{\lyap}{\mathcal{V}}

\newcommand*{\pose}[2]{\prescript{#2}{}{\bm{X}}_{#1}}
\newcommand*{\trans}[2]{\prescript{#2}{}{\bm{p}}_{#1}}

\newcommand*{\rot}[2]{\prescript{#2}{}{\bm{R}}_{#1}}

\newcommand{\textjava}[1]{{\lstset{basicstyle=\ttfamily}\lstinline@#1@}}
\newcommand{\textjavafn}[1]{{\lstset{basicstyle=\footnotesize\ttfamily}\lstinline@#1@}}
\usepackage{setspace}
\usepackage{ifthen}

\long\def\sfootnote[#1]#2{\begingroup%
\def\thefootnote{\fnsymbol{footnote}}\footnote[#1]{#2}\endgroup}
\newcommand{\SE}{\mathrm{SE}}
\newcommand{\SO}{\mathrm{SO}}
\renewcommand{\se}{\mathfrak{se}}
\newcommand{\ignore}[1]{}

\DeclareMathOperator*{\argmin}{arg\,min}

\title{\LARGE \bf
End-to-end Multi-Instance Robotic Reaching from Monocular Vision
}
\author{Zheyu Zhuang$^{1}$, Xin Yu$^{2}$, Robert Mahony$^{1}$
\thanks{This research was supported by the Australian Research Council
through the ``Australian Centre of Excellence for Robotic Vision'' CE140100016.}%
\thanks{$^1$ Zheyu Zhuang, Robert Mahony are with ``Australian Centre for Robotic Vision'', Research School of Engineering, The Australian National University. {\tt\small first.last@anu.edu.au}}
\thanks{$^2$ Xin Yu is with School of Computer Science, The University of Technology Sydney. {\tt\small first.last@uts.edu.au}}
}%

\begin{document}

\maketitle
\thispagestyle{empty}
\pagestyle{empty}

\begin{abstract}
Multi-instance scenes are especially challenging for end-to-end visuomotor (image-to-control) learning algorithms.
``Pipeline'' visual servo control algorithms use separate detection, selection and servo stages, allowing algorithms to focus on a single object instance during servo control.
End-to-end systems do not have separate detection and selection stages and need to address the visual ambiguities introduced by the presence of arbitrary number of visually identical or similar objects during servo control.
However, end-to-end schemes avoid embedding errors from detection and selection stages in the servo control behaviour, are more dynamically robust to changing scenes, and are algorithmically simpler.
In this paper, we present a real-time end-to-end visuomotor learning algorithm for multi-instance reaching.
The proposed algorithm uses a monocular RGB image and the manipulator's joint angles as the input to a light-weight fully-convolutional network (FCN) to generate control candidates.
A key innovation of the proposed method is identifying the optimal control candidate by regressing a control-Lyapunov function (cLf) value.
The multi-instance capability emerges naturally from the stability analysis associated with the cLf formulation.
We demonstrate the proposed algorithm effectively reaching and grasping objects from different categories on a table-top amid other instances and distractors from an over-the-shoulder monocular RGB camera.
 The network is able to run up to $\sim$160 fps during inference on one GTX 1080 Ti GPU.

\end{abstract}

\section{INTRODUCTION}
Scenes that contain multi-instances are common in our daily life, for example; cutlery sets on dining tables, books and stationary on desks, fruit hanging on trees, \textit{etc}.
As robotic systems transition into more real-world and shared autonomy environments, the ability to reach and grasp objects amid distracters and in the presence of visually similar objects becomes critical.
Pipeline approaches follow the  ``detect, decide, then servo" paradigm.
Grasping algorithms such as \cite{zeng2017multi, morrison2018cartman} first detect objects and estimate 6DoF poses for each detected object.
The selection module chooses which object to grasp and the estimated grasp pose is provided to the servo module.
The grasp action is undertaken using the standard pose control for robotic manipulators.
Such a pipeline can be run in real-time if the computational requirements can be met and a heuristic is used to ensure the algorithm does not jump between different grasp targets unintentionally.

Convolutional neural networks have become the default algorithm for detection and pose estimation from visual data.
Algorithms that first regress an intermediate representation, such as image keypoints, and then compute object pose by solving a PnP problem~\cite{peng2019pvnet, song2020hybridpose, zakharov2019dpod}, have achieved impressive performance on popular monocular pose estimation datasets including LINEMOD~\cite{brachmann2014learning} and Occlusion LINEMOD~\cite{hinterstoisser2012model}.
However, existing algorithms are trained on single instance datasets and it is unclear how the underlying architecture will adapt to multi-instance.
Tremblay \textit{et al.}~\cite{tremblay2018deep} propose a real-time pose estimation network with multi-instance capability and showcase repeatable experiments of robotic grasping.
To authors understanding, this work \cite{tremblay2018deep} is the state-of-the-art result in multi-instance visual reaching and grasping.
\begin{figure}[t]
	\centering
	\includegraphics[width=0.49\textwidth]{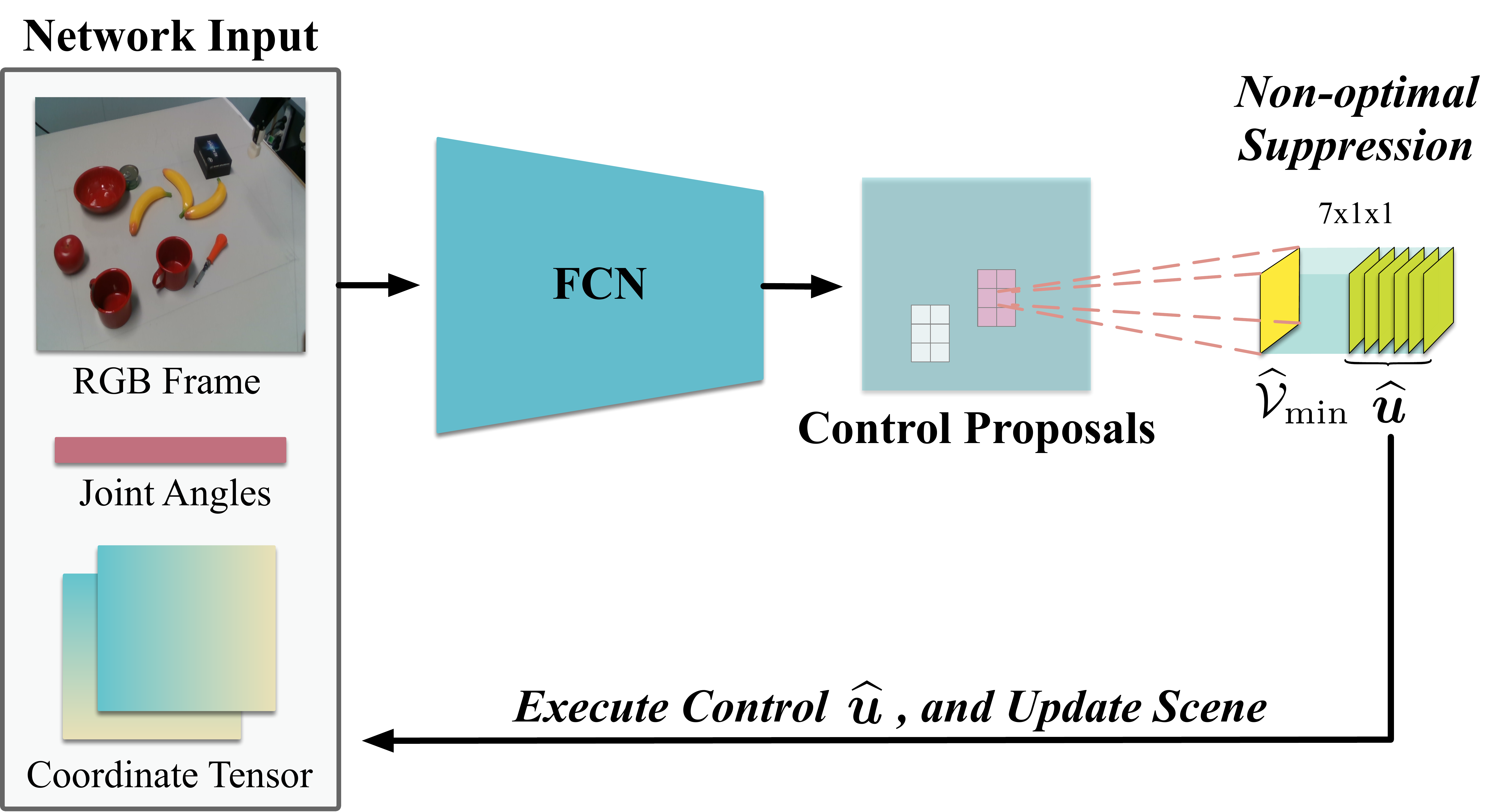}
	\caption{Architecture of the proposed closed-loop reaching algorithm.
A fully-convolutional network densely predicts a control Lyapunov function (cLf) value $\widehat{\mathcal{V}}$ and control $\widehat{u}$ associated to each foreground image grid cell.
Non-optimal suppression is achieved by selecting the control associated with the grid cell corresponding to the lowest cLf value.
The control is updated in real-time as the image and joint angles are updated.
The reaching trajectory terminates when the regressed Lyapunov value is lower than a threshold.}
	\label{fig:pipeline}
\end{figure}
While separating the computer vision pose estimation from servo control is conceptually simple, this approach can lead to undesirable error propagation between modules~\cite{tremblay2018deep}.
Furthermore, in dynamic scenes, the decision module needs to associate and track the estimated poses between frames~\cite{corke2000real}.

In contrast, end-to-end approaches are known to be robust to dynamic scene changes and model errors~\cite{hutchinson1996tutorial}.
Levine \textit{et al.}~\cite{levine2016end} demonstrate a robot accomplishing real-world tasks such as hanging clothes and screwing on a bottle cap by executing control learnt from RGB inputs.
James \textit{et al.}~\cite{james2017transferring} transfer a recurrent network that learns a multi-stage task from simulation to real-world with no real-samples.
This work shows the robot sequentially reach, grasp and place a red cube amid distracters.
Zhang~\textit{et al.}~\cite{zhang2019adversarial} achieve closed-loop reaching towards one unique target in a clutter from monocular RGB images by learning a visuomotor policy from a pose-based controller.
To increase the network's accuracy,~\cite{james2017transferring} learns the target pose from the image feature vector as an auxiliary task, and~\cite{zhang2019adversarial} pre-trains the CNN feature extractor by regressing the target pose.
These techniques apply only to the single instance scenario. For multi-instance image-to-control learning algorithms,
Zhuang \textit{et al.}~\cite{zhuang2019learning} use similar fully-connected network architecture as in~\cite{zhang2019adversarial, james2017transferring}. However, a hot-swap scheme is proposed to attend network's focus onto a single instance during the final stages of servo control.

In this paper, we propose a fully-convolutional network that densely predicts control action over a grid of image cells, associating the dominant visible instance of the desired object category with the cell.
This naturally gives the network the ability to generalise over an arbitrary number of object instances visible in different grid cells.
A key innovation is that we generate the desired control action from a control Lyapunov function formulation, and separately regress the value of the Lyapunov function alongside the corresponding control.
The algorithm then uses the regressed Lyapunov function value to select the control action associated with the lowest Lyapunov value, ensuring that the closed-loop system is drawn to the object that is easiest to grasp. This behaviour is encoded in the Lyapunov control function design, and the selected control acts to continuously decrease the Lyapunov function value, leading to successful grasp actions.

In summary, the key contributions of this paper are:
\begin{itemize}
	\item We demonstrate a real-time, closed-loop, image-to-control fully-convolutional network for robotic reaching in cluttered and dynamic environments. The proposed network achieves consistent high grasp success rate for different object categories regardless of the presence of simultaneous instances and visual clutter.
	\item We showcase the simplicity and efficiency of utilising a control Lyapunov function approach to deal with visual ambiguity associated with  multi-instance grasping.
	\item We demonstrate that the proposed approach can be trained entirely on simulated data and transfer effectively to real world scenarios.
\end{itemize}

\section{Formulation}
\label{sec:cLf_design}

This section presents the formulation of the control Lyapunov function and corresponding velocity control for a reaching task.

The 6 DoF poses of target frame $\{G\}$ and end-effector frame $\{H\}$ are represented by elements of the Special Euclidean Group $\SE(3)$.
Denote the pose of a frame $\{B\}$ with respect to a reference frame $\{A\}$ as $\pose{B}{A}$, and its rotation matrix and translation vector as $\rot{B}{A} \in \SO(3)$ and $\trans{B}{A}\in \R^3$ respectively.
The left superscript is omitted if the pose is defined with respect to the world reference frame.
The absolute end-effector pose $\pose{H}{} = \pose{H}{}(\bm{\theta})$ is a function of joint angles $\bm{\theta}\in\R^{6\times1}$, i.e., the forward kinematics model of the manipulator.

\subsection{Symmetry-aware Control Lyapunov Function}
A control Lyapunov function (cLf) for a reaching task is a continuously differentiable scalar-valued positive-definite function $\lyap(\bm{\theta})$ of the joint angles. $\lyap(\bm{\theta})$  is zero only at the joint coordinates for the desired goal pose $\{G_i\}$ of an object $i$.
We formulate the cLf as:
\begin{align}
 \lyap(\pose{H}{}) :=   \frac{1}{2}\|\pose{H}{}-\pose{G_i}{}\|^2_{kF},
	\label{eq:lyap}
\end{align}
where $\| \cdot \|_{kF}$ denotes a preferentially weighted Frobenius norm.
In particular, we preferentially weight the translation component of the pose matrix to balance the relative sensitivity of the homogeneous transform to rotational and translation displacements.
In this work, a preferential weighting of 5-to-1, translation to rotation weighting, is determined empirically.

Many objects of interest have geometric symmetries and there are a continuum of equally valid grasp poses $\pose{G_i}{} \in \mathcal{G}_i$ for an object $i$ described by a set $\mathcal{G}_i \subset \SE(3)$.
To address this we allow the pose $\pose{G_i}{}$ in \eqref{eq:lyap} to vary within the constraint set $\mathcal{G}_i$ depending on the end-effector pose $\pose{H}{}$.
That is, for a given object $i$, then $\pose{G_i}{} := \pose{G_i}{}(\pose{H}{})$ is chosen as a function of the end-effector pose
\begin{equation*}
\pose{G_i}{} :=  \argmin_{\pose{G}{} \in \mathcal{G}_i} \left(\|\pose{H}{}-\pose{G}{}\|^2_{kF} \right).
\end{equation*}

\begin{figure*}
	\centering
	\subfloat[]{
		\includegraphics[width=0.98\linewidth]{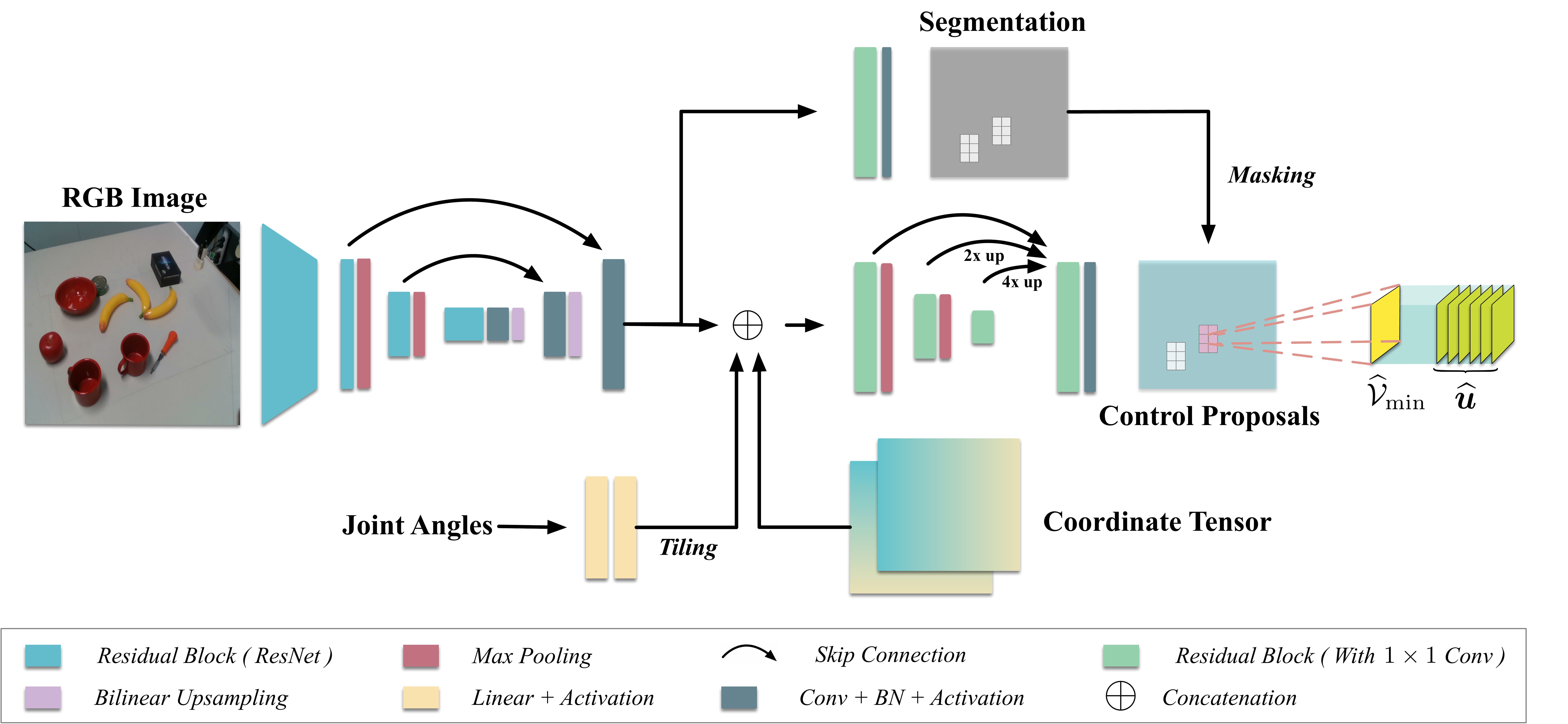}
		\label{fig:net_arch_main}}\\
	\subfloat[]{
		\includegraphics[width=0.98\textwidth]{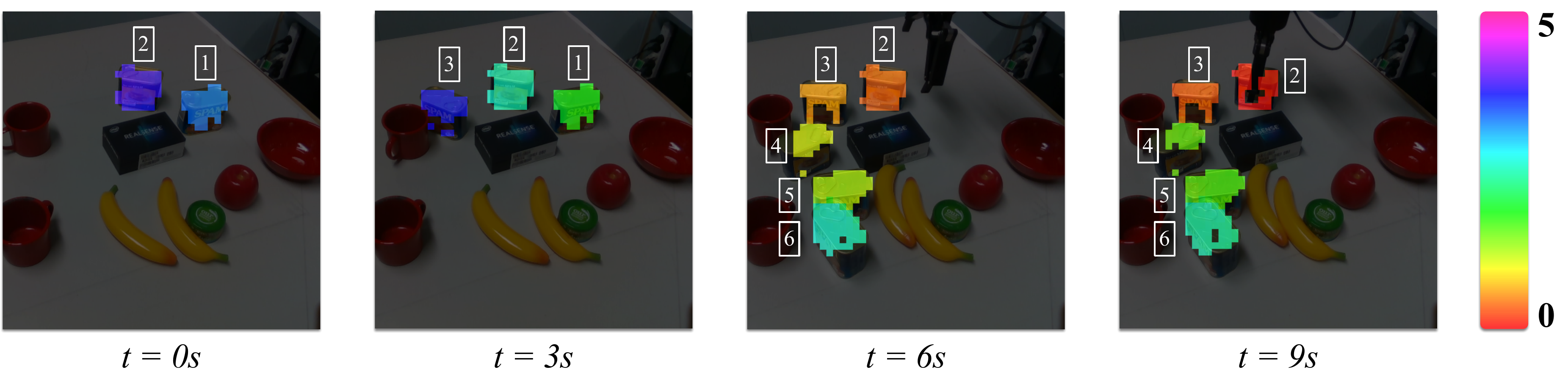}
		\label{fig:reaching_pics}}

	\caption{
		(a) The proposed network architecture.
(b) Visualisation of the dynamic robustness of the reaching performance.
The robot is undertaking a real-time reaching trajectory, however, it is stopped every 3s and the scene is rearranged.
The Lyapunov value of the image grid cell is coded by colour as shown by the colour-bar.
Initially, target instance 1 has the lowest Lyapunov value and the reaching trajectory is focusing on this instance.
The control is unchanged with the addition of another target instance 3.
After target 1 is removed, the reaching trajectory refocuses on target instance 2.
The introduction of any extra instances or distractors makes no impact on the successful grasp achieved at time $t = 9s$.}
	\label{fig:apcNet}
	\vspace{-2mm}
\end{figure*}
\subsection{Velocity Controller Design}
The velocity controller is derived based on the differentiation of the cLf to guarantee the decrease property of the cLf. This decrease property makes the proposed non-optimal suppression possible.

To formulate the velocity control, we use the velocity Jacobian $\bm{J} = \bm{J}(\bm{\theta})$ for the manipulator.
Denoting the angular and translational rigid body velocity of $\pose{H}{}$ expressed in its body-fixed frame $\{H\}$ by $\bm{\omega}\in \R^{3\times1}$ and $\bm{v}\in\R^{3\times1}$ respectively, one has $(\bm{\omega}, \bm{v})^\top = \bm{J}(\bm{\theta}) \dot{ \bm{\theta}}$.
The partial differential of cLf $\lyap$ with respect to the pose of the world frame relative to end-effector $\pose{}{H}$ frame is derived from Eq.~\eqref{eq:lyap} as:
\begin{equation*}
        \nabla{\lyap{(\pose{H}{})}} = \text{proj}^k_{\se(3)}\left(\pose{H}{}^\top(\pose{H}{}-\pose{G}{})\right) \in \se3,
\end{equation*}
where $\nabla$ is differentiation with respect to $\SE(3)$, $\text{proj}^k_{\se(3)}$ is the matrix projection operator that maps an arbitrary $4\times4$ matrix to the nearest member of $\se(3)$ measured in Frobenius norm subject to the preferential weighting $k$.
In particular, for a general matrix
\[
D = \begin{pmatrix}
D_{11} & D_{12} \\ D_{21} & D_{22}
\end{pmatrix} \in \R^{4 \times 4},
\]
with $D_{11} \in \R^{3 \times 3}$, $D_{12} \in \R^{3 \times 1}$, $D_{21} \in \R^{1 \times 3}$ and $D_{22} \in \R$,
the matrix projection is defined as:
\begin{align*}
\text{proj}^k_{\se(3)} (D) =
\begin{pmatrix}
\frac{1}{2}(D_{11} - D_{11}^\top) & k D_{12} \\
0 & 0
\end{pmatrix} \in \se(3),
\end{align*}
where $k > 0$ is the scaling factor that weights the translation control sensitivity relative to the rotational sensitivity.
Let $(\cdot)^{\vee}$ denote the linear readout mapping that takes a matrix in $\se(3)$ and forms the associated $(\bm{\omega}, \bm{v}) \in \R^6$ vector of angular and linear velocities.
The proposed joint velocity control is
\begin{equation}
    \label{eq:vel_ctrl}
    \bm{u} := -\bm{J}(\bm{\theta})^{-1}(\nabla{\lyap{}}(\pose{H}{}))^\vee,
\end{equation}
where $\bm{\theta}$ are joint angles associated to $\pose{H}{}$.
This velocity controller design guarantees, in a closed-loop system, executing the velocity control continuously decreases the Lyapunov value, i.e.
\begin{equation*}
    \dot{\lyap}  = -\text{tr}(\nabla{\lyap{}}^\top\nabla{\lyap{}})
    = -\|\nabla{\lyap{}}\|^2_{kF}<0,
\end{equation*}
where $\dot{\lyap}$ denotes the time derivative of $\lyap$.

\section{Learning the Control Lyapunov Function}
The proposed network divides the input image into square grid cells. Each cell predicts a binary foreground visibility score, a Lyapunov value and a control vector.  All grid cells corresponding to the same visible instance share identical supervision labels. In this work, the resolution of the grid cell is set to $1/8$ of the input image size.

\label{sec:III:learning}
\subsection{Network Architecture}
The network architecture is illustrated in Fig.~\ref{fig:net_arch_main}. We adopt an \textbf{image feature extractor} as the backbone of this network; the extracted image feature is shared between the two output regressors.
The architecture is built upon ResNet18~\cite{he2016deep} by an auto-encoder with skip connections.
The \textbf{Joint auto-encoder} comprises two fully-connected layers. It learns a higher dimensional latent representation of the joint angles $\bm{\theta}\in\R^{6\times1}$. 
The latent joint representation is then tiled over the spatial dimension and concatenated with image features and the coordinate tensor.
The \textbf{segmentation regressor} performs binary classification to separate the foreground from the background, and the \textbf{control regressor} infers the cLf value $\widehat{\lyap}$ and velocity control $\widehat{\bm{u}}$.
Residual block with $1\times1$ convolution in~\cite{newell2016stacked} is used in two output regressors.

The spatial invariance property of convolutional kernels grants fully-convolutional architectures the potential of generalising to an arbitrary number of object instances. 
However, for an image-to-control task, the network needs access to precise spatial information. 
Inspired by ``CoordConv" in~\cite{liu2018intriguing, wang2019solo}, 
we introduce a \textbf{coordinate tensor} input.
This is a two-channel tensor, that carries the normalised UV coordinates of each grid cells. 
Coordinate tensor is concatenated with the joint features and image features as the input to the control regressor.

\subsection{Loss Functions}
Binary cross entropy loss is used for the segmentation branch:
\begin{equation*}
    \label{eq:loss_seg}
    \C{L}_{\text{seg}} = \frac{1}{N}\sum_{n=1}^{N}\left(y_n\,\text{log}S(\widehat{\mathcal{C}}_n)+(1-y_n)\,\text{log}(1-S(\widehat{\mathcal{C}}_n))\right),\\
\end{equation*}
where $n\in\{1, \dots, N\}$ denotes
 grid cell index, $S(\cdot)$, $\widehat{\mathcal{C}}$ and $y$ represent the sigmoid function, segmentation regressor output and binary visibility label respectively.

A weighted L1 loss $\C{L}_{\text{ctrl}}$ is used for the control branch and this should not be confused with the Lyapunov function $\lyap$.
Only the loss generated by foreground grid cells are penalised. 
The control regressor's loss function is formulated as:
\begin{equation}
    \label{eq:loss_reg}
    \C{L}_{\text{ctrl}} = \frac{1}{N_\text{pos}}\sum_{n=1}^{N}y_n \left(\left|\lyap_n-\widehat{\lyap}_n\right| +
    \frac{1}{6}\left|\bm{u}_n-\widehat{\bm{u}}_n\right|\right),
\end{equation}
where ${N_\text{pos}}$ represents the total number of grid cells in the foreground.
The final loss function is: $\mathcal{L} = \mathcal{L}_\text{seg}+\mathcal{L}_\text{ctrl}.
$

\subsection{Non-optimal Suppression}
Our network predicts a tuple of values $(\widehat{\lyap},~\widehat{\bm{u}})$  at each grid cell associated with targets present in that grid cell.
An essential step in the algorithm is to select the single input $\widehat{\bm{u}}$ that is the optimal control action proposal for the robot to act on.
In the pipeline algorithms, this control choice is determined by the instance the algorithm is targeting to grasp, a decision process that must be explicitly coded. 
For end-to-end methods such as our system, this decision process emerges from the dynamic behaviour of the system. 

Since we have derived a control action from a global control Lyapunov function formulation and regress the actual value of the Lyapunov function along with the control proposal, these values provide an ideal metric of optimality for choosing the control action.
By executing the action associated with the minimal Lyapunov value chosen across the segmentation masks, the control proposal is adapted to maximally decrease that particular instance of the cLf. 
It follows, that the closed-loop motion of the robot reinforces the initial preference for a given instance. This motion decreases the cLf value associated to this instance more quickly than the cLf value associated with other instances. 
This increases the likelihood that the same instance will generate the minimal Lyapunov regressor for the next image input.
Indeed, in the absence of errors and for static scenes, the choice of instance is locked in by the initial minimisation.

For real-world dynamic scenarios, the closed-loop motion of the robot can be seen as an analogue solver for a stochastic gradient descent algorithm that both selects an instance to target and then computes the grasp pose.
We refer to this process as non-optimal suppression since the evolution of the closed-loop acts to increase confidence in the chosen instance while the closed-loop trajectory is tracked.
The progressive nature of the decision process and its integration into the servo task makes the formulation naturally robust to dynamic variations in the scene. 
A  reaching example with visualised non-optimal suppression process is shown in Fig.~\ref{fig:reaching_pics}.

Image grid cells associated with the same object instance share identical regression targets. 
In practice, it is nearly impossible for these cells to produce identical values. 
Selecting the lowest Lyapunov value amongst those cells in a given instance segmentation provides the proposed method to suppressing similar control action proposals.
Statical methods like clustering and averaging may increase the accuracy of the inference.
However, a frame-by-frame clustering and averaging algorithm adds significant computation overhead. 
Thus, we do use a cross-frame exponentially weighted moving averaging term, known as momentum in machine learning literature, to filter sudden changes in the closed-loop control action inference. 
The momentum term is define as:
\begin{equation}
\bar{\bm{u}}_t = \eta\bar{\bm{u}}_{t-1} + (1-\eta)\widehat{\bm{u}}_t,
	\label{eq:momentum}
\end{equation}
where $\eta$ is a tuneable constant $\in[0, 1]$ and $\widehat{\bm{u}}_t$ is a current raw velocity control prediction from the network. 
This is equivalent to applying a low pass filter to the raw control signal generated by the network. 
The constant $\eta$ is set to 0.5 in our experiments. 
It leads to smother reaching trajectories especially under high controller gain.

\section{Implementation}
\begin{figure}[t]
	\centering
	\subfloat[Simulation]{
		\includegraphics[width=0.20\textwidth]{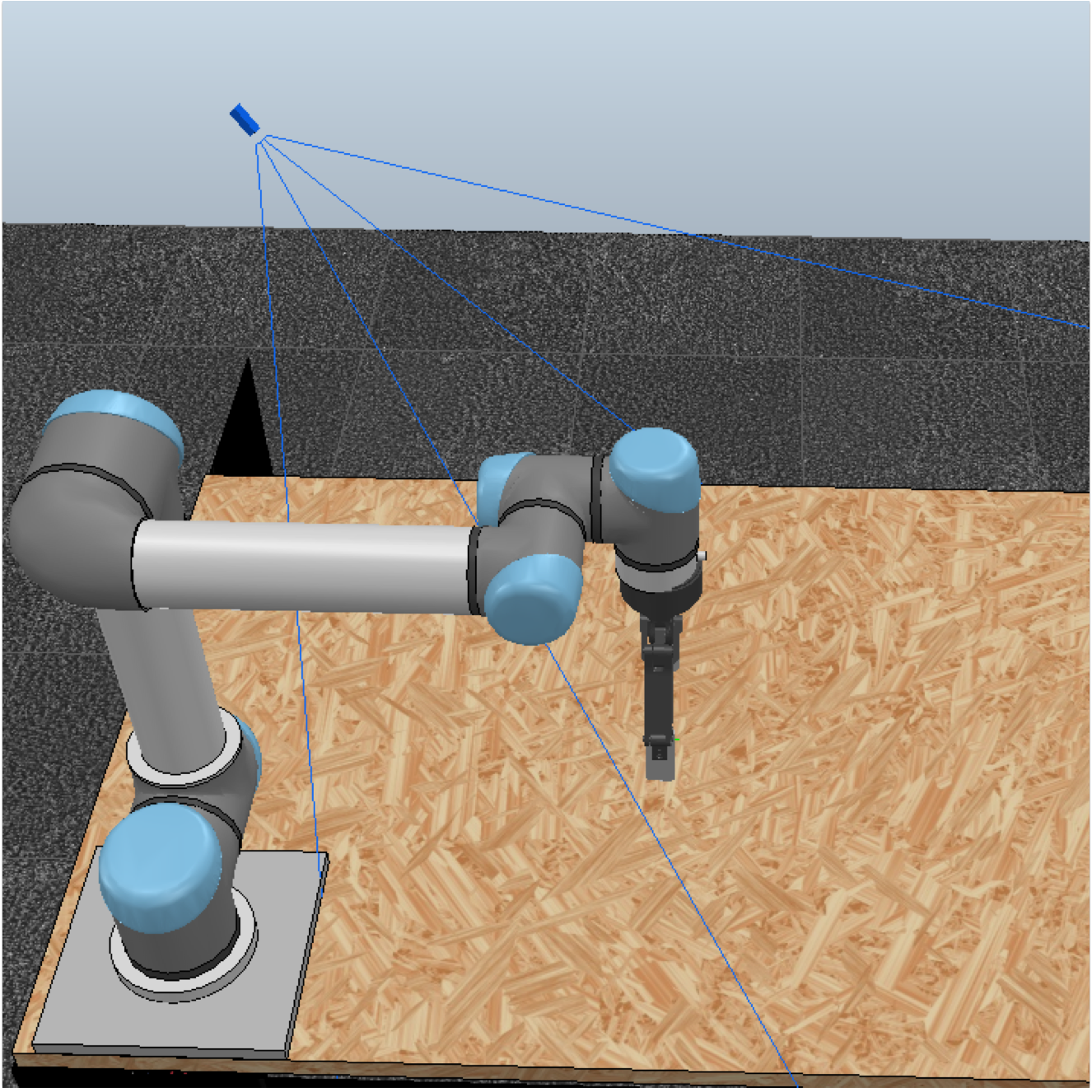}
		\label{fig:lab_setup:sim}}
	\hspace{3mm}
	\subfloat[Real-world]{
		\includegraphics[width=0.20\textwidth]{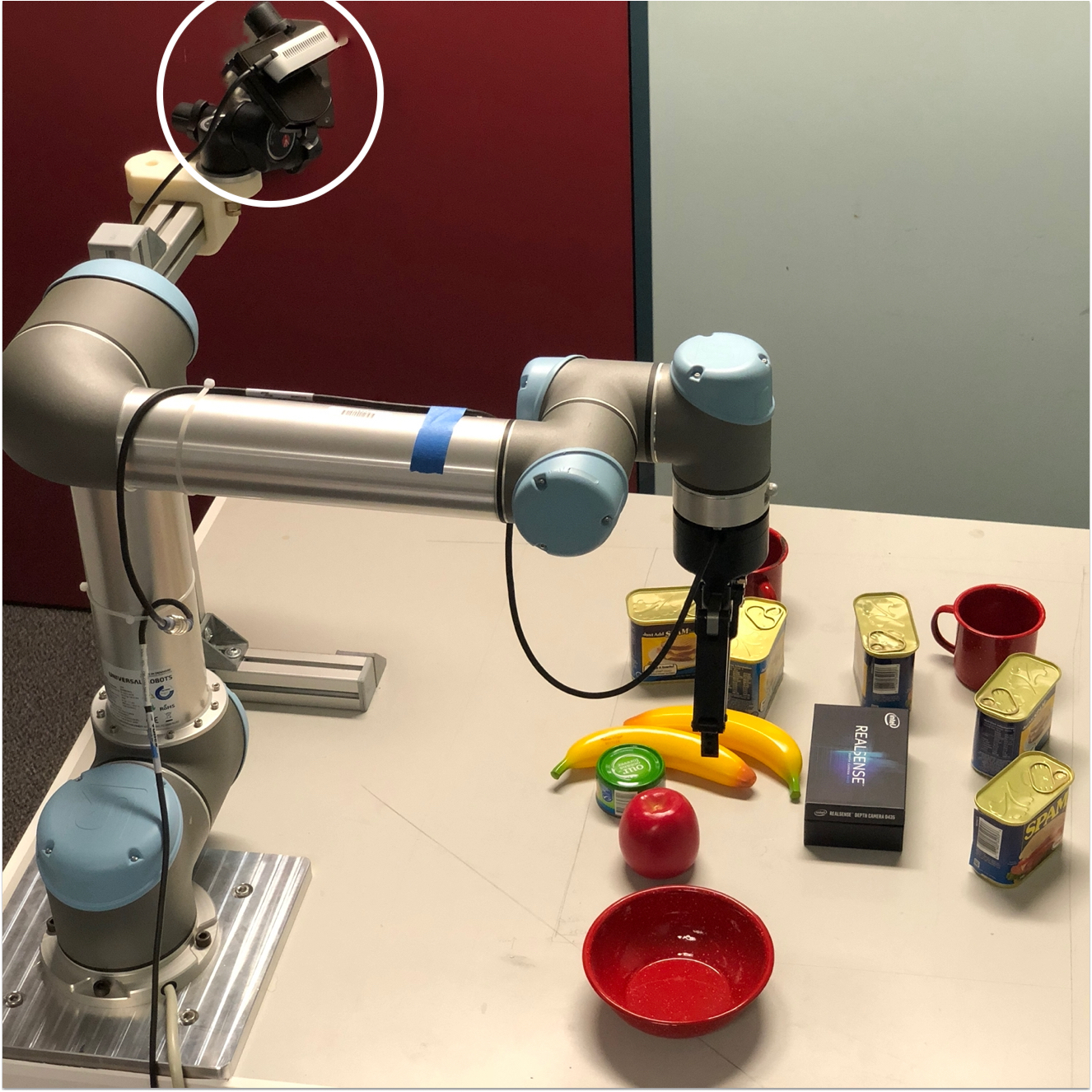}
		\label{fig:lab_setup:real}}
\caption{Lab and simulation environments: 
The first-person camera is positioned as shown in Fig.~\ref{fig:lab_setup:real} (marked with the white circle), pointing towards the table workspace.
The simulated environment is geometrically identical to the physical layout in the lab. 
The simulated camera is calibrated to simulate the real camera.}
\label{fig:lab_setup}
\vspace{-3mm}
\end{figure}

We perform robotic grasping as the ultimate test for the proposed reaching algorithm. A two-finger parallel gripper is attached to a UR5 manipulator. The finger tips of the parallel gripper are padded with textured soft-silicone to increase the contact area. A Realsense D435 is place over the shoulder of the manipulator (see Fig.\ref{fig:lab_setup}). Only the RGB camera of Realsense is used for the experiments. The camera's frame rate is set to 60Hz and the resolution is resized to $512\times384$.

\label{sec:implementation}

\subsection{Data Collection}
\label{subsec:simulator}
Generating large-scale, pre-labeled data with simulators is almost cost-free compared to collecting the real-world data.
Deploying a network trained purely on synthetic data to the real-world has been proven feasible~\cite{tremblay2018deep, james2017transferring}.
In this work, we replicate the real-world setting in Coppeliasim.
The camera extrinsics are calibrated to the manipulator base by observing a checkerboard attached to the end-effector.

For each sample, the number of simultaneous instances is equally sampled between one and three, and instances are simulated on the tabletop amongst a collection of random distractors.
The end-effector is positioned at a random 6 DoF pose within the workspace of the robot. 
The end-effector's initial simulated translation component is sampled on the surface of a quarter sphere, whose radius is sampled from uniform distribution and constrained by the manipulator's usable workspace.
The  initial simulated rotation is sampled based on axis-angle representation; the direction axis is sample inside a downward-facing cone, and the angle is sample in $[-\pi,~\pi]$. 
Domain randomisation~\cite{tobin2017domain} is only applied to all visible, non-object entities. 
In order to increase data efficiency and  improve the local convergence of the closed-loop system, we increase the sampling density while the manipulator approaches the neighbourhood of a target.

In this work, we collect three datasets for mugs, IKEA LACK Table Legs (table leg for short), and potted meat (Spam) cans. 
Mug and Spam are from the YCB dataset~\cite{calli2015ycb}. 
The ``multi-table leg" dataset contains up to two simultaneous instances; the  ``multi-spam" and ``multi-mug" dataset contain up to three instances. 
No spam was consumed in the development of this paper. 


\subsection{Network Training Details}
\label{subsec:learn_results}
Our training dataset contains approximately 55k simulated samples for each object category.
We train our networks with 90\% of the training dataset while using the remaining 10\% for evaluation.
The brightness, saturation, contrast and hue of input images are randomly jittered at 10\% of their maximum ranges to alleviate the domain gap between simulated and real data.
The weights of ResNet18 backbone are randomly initialised. 
We use ADAM optimiser with the batch size 64 for learning. 
The learning rate is initialised as $10^{-3}$ with a decay rate 0.8 for every 5 epochs. 
The maximum training epoch is set to 80. 

\begin{figure*}
	\centering
	\subfloat[With CoordConv]{
		\includegraphics[width=0.45\textwidth]{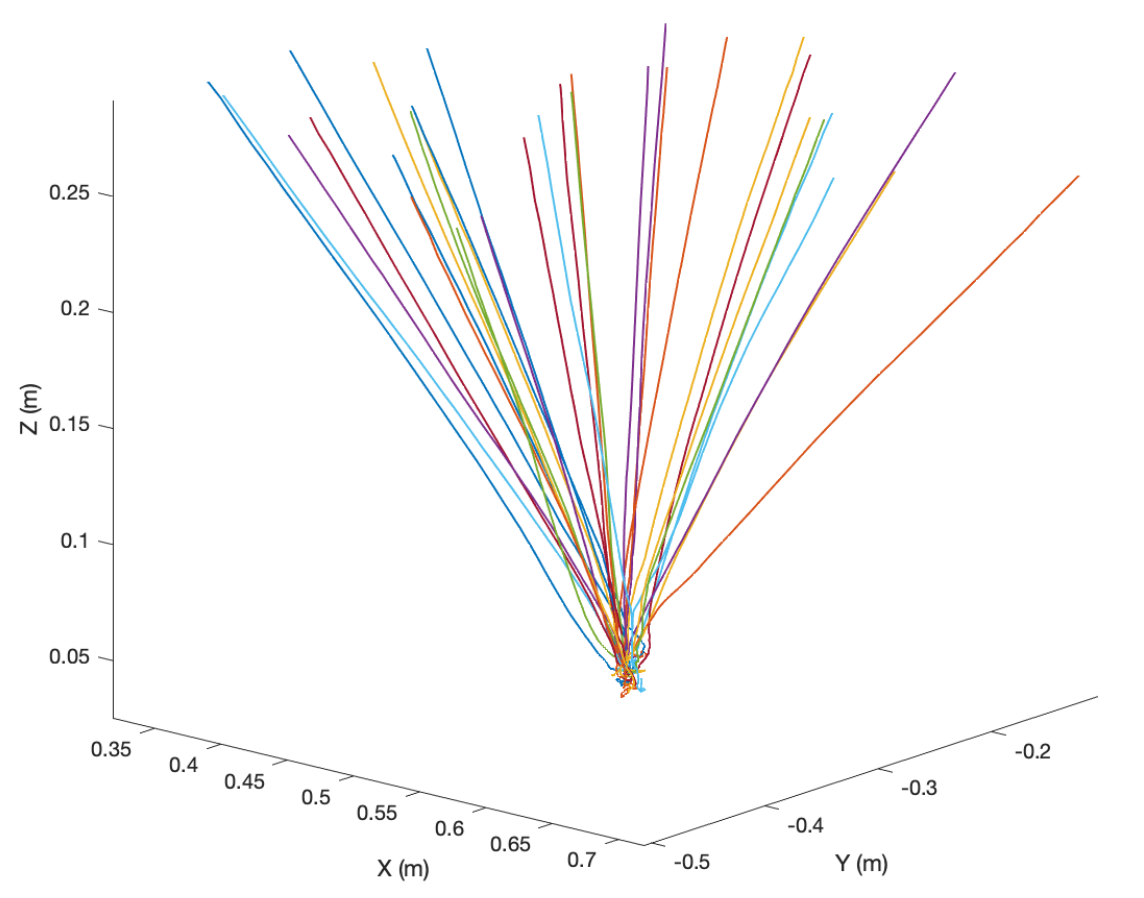}
				\label{fig:coord_conv_traj}}
	\hspace{3mm}
	\subfloat[Without CoordConv]{
		\includegraphics[width=0.45\textwidth]{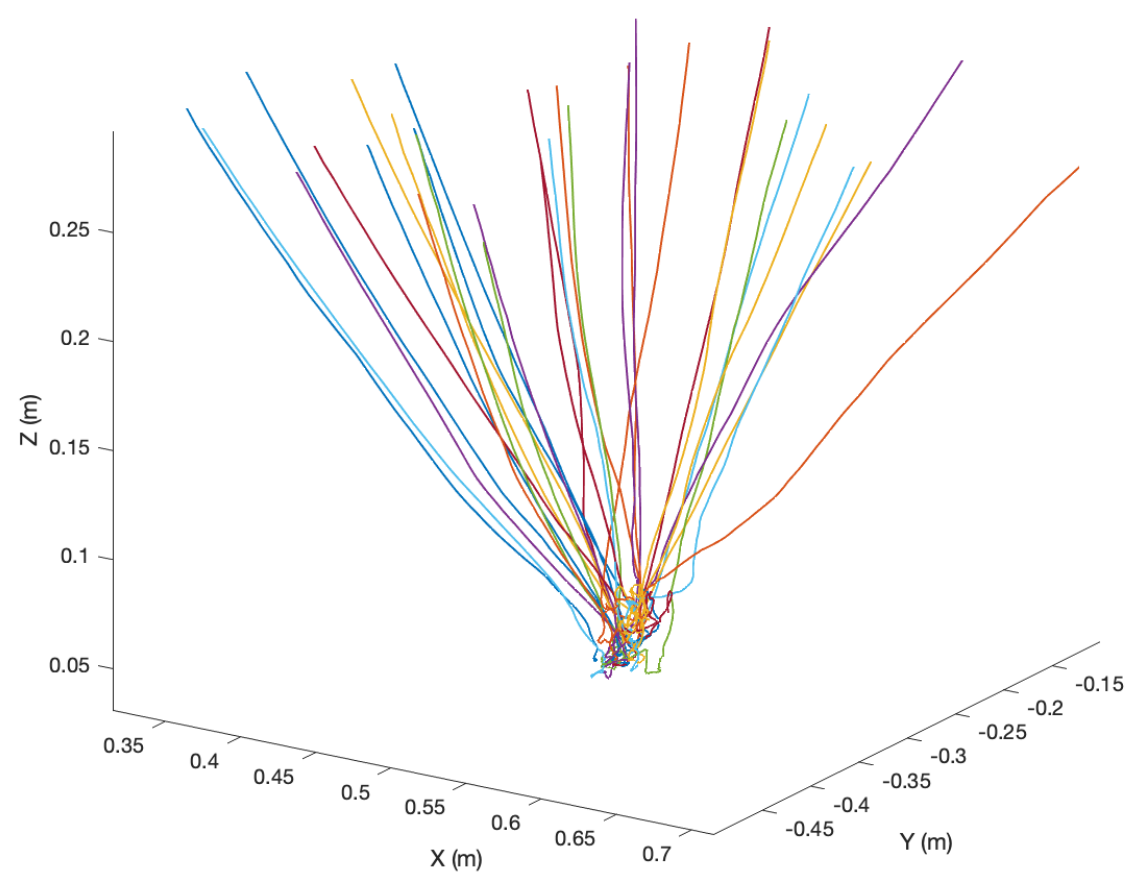}
	\label{fig:no_coord_conv_traj}}

	\caption{
Visualisation of 32 reaching trajectories for two networks trained with and without CoordConv for the ``multi-spam'' dataset.
For a fair comparison, experiments with each network share the same pre-sampled random initial end-effector poses. 
The test scene is static, and contains one Spam Can and distractors.
The vertical axis indicates the control regression loss $\mathcal{L}_\text{ctrl}$ defined in Eq.~\eqref{eq:loss_reg} .}
\vspace{-3mm}
\end{figure*}

\begin{table}
	\begin{center}
		\begin{tabular}{c c c c c c c}
			\toprule
			& \textbf{\makecell{Coord-\\Conv}} & \multicolumn{5}{c}{\textbf{\makecell{Grasp\\ Success}}}\\
			\midrule
			 &  & \small 1 & \small 2 & \small 3 & \small 4 & $\%$\\
			\midrule
			\multirow{2}{*}{\textbf{Spam}}
			&\small with & \textbf{13} & \textbf{14} & \textbf{15} & \textbf{14} & \textbf{93.3}\\
			& \small w/o & 12 & 11 & 12 & 11 & 76.7\\
			\midrule
			\multirow{2}{*}{\textbf{\makecell{LACK \\Leg}}}
			&\small with & \textbf{14} & \textbf{14} & \textbf{14} & \textbf{12} & \textbf{90.0}\\
			& \small w/o & 9 & 10 & 9 & 9 & 61.7 \\
			\midrule
			\multirow{2}{*}{\textbf{Mug}}
			&\small with & \textbf{14} & \textbf{14} & \textbf{15} & \textbf{14} & \textbf{95.0}\\
			& \small w/o & 13 & 13 & 13 & 14 & 88.3 \\
			\midrule
			\multirow{2}{*}{\textbf{Avg. (\%)}}
			& \small with & - & - & - & - & \textbf{92.8} \\
			& \small w/o & - & - & - & - & 75.6\\
			\bottomrule
		\end{tabular}
	\end{center}
	\caption{Statistical results of real-world grasping experiments using with and without CoordConv. 60 reaching experiments (15 for one, two, three and four simultaneous instances) are conducted for each object category.}
	\label{tab:main_results}
\vspace{-3mm}
\end{table}

\section{Grasping Experiments}

In this section, we demonstrate the performance and robustness of the proposed reaching method through multi-instance real-world grasping experiments. Mugs, Table Legs, and Spam cans are used as testing objects.

For each object category, we separately conduct 60 reaching experiments (15 for one, two, three and four simultaneous instances). 
The gripper closes when the Lyapunov value is below a designated threshold. We set this threshold to 0.005 for all our experiments.
In each scene, the end-effector pose is randomly initialised in the workspace, and target instances and distractors are shuffled. 
IKEA table legs have larger grasping error tolerance due to their elongated geometric shape. Hence, for IKEA table legs, we only register a successful grasp when the gripper closes within $\pm3~cm$ from the designated grasp point.
The grasping strategy for mugs is designed as opening the gripper when its tip is inside the cavity of mugs.
That is we grasp from the inside out for mugs. 

As shown in Tab.\ref{tab:main_results}, the proposed method achieves on average $92.8\%$ grasp success rate over 3 different object categories without leveraging additional sim-to-real transfer techniques apart from domain randomisation. 
More importantly, the proposed algorithm maintains its high accuracy when an arbitrary number of simultaneous instances are present.
The network also generalises beyond the number of instances included in the training dataset for three object categories.
The proposed algorithm exhibits strong false positive rejection capability.
Most of false positives in the vision pipeline correspond to Lyapunov values larger than the current minimum and are automatically rejected by our non-optimal suppression algorithm.
Occasional non-persistent false positives with lower Lyapunov values do disrupt the control, however, this effect is not significant and is strongly mitigated by the momentum term added to the closed-loop controller design formulated in Eq.\eqref{eq:momentum}.

The proposed method achieves comparable overall grasp success rate to the state-of-the-art pose estimation based multi-instance grasping~\cite{tremblay2018deep}. 
Note that, in our experiments the degrees of freedom of the object poses are constrained since they lie on the table. 
However, our system regresses a full 6 DoF control and the relative pose of the initial end-effector with respect to the targets are unconstrained.
A further key point is that the proposed network can run at up to 160 fps on a single GTX 1080 Ti GPU. 
The camera that we use only runs at 60Hz reducing the control bandwidth used the experimental studies reported, however, the computational complexity of the approach is not a limiting factor for higher fps than 60Hz. 
The real-time computational performance of the proposed algorithm is significantly ahead of state-of-the-art methods that use pipeline architectures (approximately 10 fps in~\cite{tremblay2018deep}). 

\section{Ablation Study}

The \textbf{CoordConv} module that tiles the grid cell representation with coordinates is a crucial component of the proposed network, enhancing stable and reliable real-world reaching.
Islam \textit{et al.}~\cite{islam2019much} provide evidence to support that spatial information can be implicitly learnt by applying zero-padding operations.
However, this implicitly learnt spatial information has proved to be insufficient for accurate visuomotor reaching task.
Including the coordinate labels into the input to the following convolutional layers provides a simple way to compute geometric information from the image. 
 
We observe the two networks with and without CoordConv have nearly identical training and evaluation convergence for the three synthetic datasets.
However, in the real-world, without the CoordConv module, grasping success rate drops from 92.8\% to 75.6\% (Tab.\ref{tab:main_results}). 
The major cause of failures is the imprecise and unstable final reaching trajectory convergence. 
That is, the network without the CoordConv module is able to identify targets, roughly servo control towards the goal, but lacks more accurate geometric information for reliable grasping. 

For further investigation, we additionally perform 32 sets of reaching experiments on one static scene with a single instance. 
In each set of experiments, the networks with and without CoordConv separately servo the robot from identical initial end-effector pose to the target instance. The end-effector pose is randomised among different sets. Trajectories of tool-centre point are visualised in Fig.\ref{fig:coord_conv_traj} and Fig.\ref{fig:no_coord_conv_traj}. 
It is clear that the network with CoordConv produces significantly smoother trajectories and more stable and precise local reaching convergence.

The \textbf{joint auto-encoder} is another important source of information for the network regression. 
This allows the network to learn a higher dimensional latent representation of lower dimensional input (i.e. joint angles in $\R^6$) prior to being concatenated with image feature and coordinate tensor.
Although two networks with and without the joint auto-encoder exhibit similar training convergence on three synthetic datasets, the inclusion of joint auto-encoder achieves better generalisation over the evaluation datasets; approximately 47.9\% 30.6\% and 6.0\% relative performance improvements are observed on the ``multi-spam", ``multi-table leg", and``multi-mug" datasets respectively. 

\section{Conclusion}

We propose a fully-convolutional, image-to-control network for a multi-instance robotic reaching task.
In particular, we formulate control actions based on a control Lyapunov function and regress both the proposed control and the associated Lyapunov value output at a grid cell level in an end-to-end manner. 
This provides us a natural structure to implement the proposed effective and efficient non-optimal suppression strategy and reach and grasp robustly in highly complex dynamic scenes.  
Our system is trained entirely on synthetic data yet robust against the sim-to-real domain gap. 
Real-world experiments on three different object categories demonstrate the system is able to reach and with high accuracy of 92.8\% amid distractors and an arbitrary number of object instances. 

\bibliographystyle{IEEEtran}
\bibliography{main}

\begin{thebibliography}{10}
\providecommand{\url}[1]{#1}
\csname url@rmstyle\endcsname
\providecommand{\newblock}{\relax}
\providecommand{\bibinfo}[2]{#2}
\providecommand\BIBentrySTDinterwordspacing{\spaceskip=0pt\relax}
\providecommand\BIBentryALTinterwordstretchfactor{4}
\providecommand\BIBentryALTinterwordspacing{\spaceskip=\fontdimen2\font plus
\BIBentryALTinterwordstretchfactor\fontdimen3\font minus \fontdimen4\font\relax}
\providecommand\BIBforeignlanguage[2]{{%
\expandafter\ifx\csname l@#1\endcsname\relax
\typeout{** WARNING: IEEEtran.bst: No hyphenation pattern has been}%
\typeout{** loaded for the language `#1'. Using the pattern for}%
\typeout{** the default language instead.}%
\else
\language=\csname l@#1\endcsname
\fi
#2}}

\bibitem{zeng2017multi}
A.~Zeng, K.-T. Yu, S.~Song, D.~Suo, E.~Walker, A.~Rodriguez, and J.~Xiao, ``Multi-view self-supervised deep learning for 6d pose estimation in the amazon picking challenge,'' in \emph{2017 IEEE international conference on robotics and automation (ICRA)}.\hskip 1em plus 0.5em minus 0.4em\relax IEEE, 2017, pp. 1386--1383.

\bibitem{morrison2018cartman}
D.~Morrison, A.~W. Tow, M.~Mctaggart, R.~Smith, N.~Kelly-Boxall, S.~Wade-Mccue, J.~Erskine, R.~Grinover, A.~Gurman, T.~Hunn, \emph{et~al.}, ``Cartman: The low-cost cartesian manipulator that won the amazon robotics challenge,'' in \emph{2018 IEEE International Conference on Robotics and Automation (ICRA)}.\hskip 1em plus 0.5em minus 0.4em\relax IEEE, 2018, pp. 7757--7764.

\bibitem{peng2019pvnet}
S.~Peng, Y.~Liu, Q.~Huang, X.~Zhou, and H.~Bao, ``Pvnet: Pixel-wise voting network for 6dof pose estimation,'' in \emph{Proceedings of the IEEE Conference on Computer Vision and Pattern Recognition}, 2019, pp. 4561--4570.

\bibitem{song2020hybridpose}
C.~Song, J.~Song, and Q.~Huang, ``Hybridpose: 6d object pose estimation under hybrid representations,'' in \emph{Proceedings of the IEEE/CVF Conference on Computer Vision and Pattern Recognition}, 2020, pp. 431--440.

\bibitem{zakharov2019dpod}
S.~Zakharov, I.~Shugurov, and S.~Ilic, ``Dpod: 6d pose object detector and refiner,'' in \emph{Proceedings of the IEEE International Conference on Computer Vision}, 2019, pp. 1941--1950.

\bibitem{brachmann2014learning}
E.~Brachmann, A.~Krull, F.~Michel, S.~Gumhold, J.~Shotton, and C.~Rother, ``Learning 6d object pose estimation using 3d object coordinates,'' in \emph{European conference on computer vision}.\hskip 1em plus 0.5em minus 0.4em\relax Springer, 2014, pp. 536--551.

\bibitem{hinterstoisser2012model}
S.~Hinterstoisser, V.~Lepetit, S.~Ilic, S.~Holzer, G.~Bradski, K.~Konolige, and N.~Navab, ``Model based training, detection and pose estimation of texture-less 3d objects in heavily cluttered scenes,'' in \emph{Asian conference on computer vision}.\hskip 1em plus 0.5em minus 0.4em\relax Springer, 2012, pp. 548--562.

\bibitem{tremblay2018deep}
J.~Tremblay, T.~To, B.~Sundaralingam, Y.~Xiang, D.~Fox, and S.~Birchfield, ``Deep object pose estimation for semantic robotic grasping of household objects,'' in \emph{Conference on Robot Learning}, 2018, pp. 306--316.

\bibitem{corke2000real}
P.~I. Corke and S.~A. Hutchinson, ``Real-time vision, tracking and control,'' in \emph{Proceedings 2000 ICRA. Millennium Conference. IEEE International Conference on Robotics and Automation. Symposia Proceedings (Cat. No. 00CH37065)}, vol.~1.\hskip 1em plus 0.5em minus 0.4em\relax IEEE, 2000, pp. 622--629.

\bibitem{hutchinson1996tutorial}
S.~Hutchinson, G.~D. Hager, and P.~I. Corke, ``A tutorial on visual servo control,'' \emph{IEEE transactions on robotics and automation}, vol.~12, no.~5, pp. 651--670, 1996.

\bibitem{levine2016end}
S.~Levine, C.~Finn, T.~Darrell, and P.~Abbeel, ``End-to-end training of deep visuomotor policies,'' \emph{The Journal of Machine Learning Research}, vol.~17, no.~1, pp. 1334--1373, 2016.

\bibitem{james2017transferring}
S.~James, A.~J. Davison, and E.~Johns, ``Transferring end-to-end visuomotor control from simulation to real world for a multi-stage task,'' in \emph{Conference on Robot Learning}, 2017, pp. 334--343.

\bibitem{zhang2019adversarial}
F.~Zhang, J.~Leitner, Z.~Ge, M.~Milford, and P.~Corke, ``Adversarial discriminative sim-to-real transfer of visuo-motor policies,'' \emph{The International Journal of Robotics Research}, vol.~38, no. 10-11, pp. 1229--1245, 2019.

\bibitem{zhuang2019learning}
Z.~Zhuang, J.~Leitner, and R.~Mahony, ``Learning real-time closed loop robotic reaching from monocular vision by exploiting a control lyapunov function structure,'' in \emph{2019 IEEE/RSJ International Conference on Intelligent Robots and Systems (IROS)}.\hskip 1em plus 0.5em minus 0.4em\relax IEEE, 2019, pp. 4752--4759.

\bibitem{he2016deep}
K.~He, X.~Zhang, S.~Ren, and J.~Sun, ``Deep residual learning for image recognition,'' in \emph{Proceedings of the IEEE conference on computer vision and pattern recognition}, 2016, pp. 770--778.

\bibitem{newell2016stacked}
A.~Newell, K.~Yang, and J.~Deng, ``Stacked hourglass networks for human pose estimation,'' in \emph{European conference on computer vision}.\hskip 1em plus 0.5em minus 0.4em\relax Springer, 2016, pp. 483--499.

\bibitem{liu2018intriguing}
R.~Liu, J.~Lehman, P.~Molino, F.~P. Such, E.~Frank, A.~Sergeev, and J.~Yosinski, ``An intriguing failing of convolutional neural networks and the coordconv solution,'' in \emph{Advances in Neural Information Processing Systems}, 2018, pp. 9605--9616.

\bibitem{wang2019solo}
X.~Wang, T.~Kong, C.~Shen, Y.~Jiang, and L.~Li, ``Solo: Segmenting objects by locations,'' \emph{arXiv preprint arXiv:1912.04488}, 2019.

\bibitem{tobin2017domain}
J.~Tobin, R.~Fong, A.~Ray, J.~Schneider, W.~Zaremba, and P.~Abbeel, ``Domain randomization for transferring deep neural networks from simulation to the real world,'' in \emph{2017 IEEE/RSJ International Conference on Intelligent Robots and Systems (IROS)}.\hskip 1em plus 0.5em minus 0.4em\relax IEEE, 2017, pp. 23--30.

\bibitem{calli2015ycb}
B.~Calli, A.~Singh, A.~Walsman, S.~Srinivasa, P.~Abbeel, and A.~M. Dollar, ``The ycb object and model set: Towards common benchmarks for manipulation research,'' in \emph{2015 international conference on advanced robotics (ICAR)}.\hskip 1em plus 0.5em minus 0.4em\relax IEEE, 2015, pp. 510--517.

\bibitem{islam2019much}
M.~A. Islam, S.~Jia, and N.~D. Bruce, ``How much position information do convolutional neural networks encode?'' in \emph{International Conference on Learning Representations}, 2019.

\end{thebibliography}

\end{document}